\documentclass[journal]{IEEEtran}
\ifCLASSINFOpdf
\else
\fi
\usepackage[cmex10]{amsmath}
\usepackage{mathtools}
\usepackage{amsfonts} 
\usepackage{siunitx}
\usepackage{algorithm,algorithmic}
\usepackage{booktabs}
\usepackage{threeparttable}
\usepackage{mathrsfs,amsmath}
\usepackage{multirow}
\usepackage{makecell}
\usepackage{placeins}
\usepackage{fixltx2e}
\usepackage{color}

\usepackage{color}

\usepackage{multirow}

\usepackage[numbers,sort&compress]{natbib}

\DeclareSIUnit\Molar{\textsc{m}} 
\DeclareSIUnit{\pH}{pH}
\DeclareSIUnit{\pixel}{px}

\renewcommand{\vec}[1]{\text{\boldmath$#1$}} 

\renewcommand{\vec}[1]{\text{\boldmath$#1$}} 

\DeclareMathOperator*{\argminA}{arg\,min} 
\DeclareMathOperator*{\argmaxA}{arg\,max} 

\newcommand\norm[1]{\left\lVert#1\right\rVert}

\begin{document}

\title{Deep Adversarial Training for Multi-Organ Nuclei Segmentation in Histopathology Images}

\author{Faisal Mahmood, Daniel Borders, Richard Chen, Gregory N. McKay, Kevan J. Salimian, Alexander Baras, \\and Nicholas J. Durr
\thanks{F.M. D.B. R.C. G.M. N.J.D.  are with the Department of Biomedical Engineering, Johns Hopkins University (JHU), Baltimore, MD 21218 and K.S. and A.B. is with the Department of Pathology, Johns Hopkins Hospital, Baltimore, MD 21287. \{faisalm, ndurr\}@jhu.edu}
\thanks{Images are best viewed in color on the electronic version of this document.}}


\maketitle

\begin{abstract}

Nuclei segmentation is a fundamental task for various computational pathology applications, including nuclei morphology analysis, cell type classification, and cancer grading. Deep learning has emerged as a powerful approach to segment nuclei, but the accuracy of convolutional neural networks (CNNs) depends on the volume and quality of labeled histopathology data for training. Moreover, conventional CNN-based approaches struggle to distinguish overlapping and clumped nuclei because they lack the capability for structured prediction. Here, we present an approach to nuclei segmentation that overcomes these challenges by utilizing a conditional generative adversarial network (cGAN) trained with synthetic and real data. We generate a large dataset of H\&E training images with perfect nuclei segmentation labels using an unpaired GAN framework. This synthetic data along with real histopathology data from different organs are used to train a conditional GAN with spectral normalization and gradient penalty for nuclei segmentation. This adversarial regression framework enforces higher-order consistency when compared to conventional CNN models. We demonstrate that this nuclei segmentation approach generalizes across different organs, sites, patients and disease states, and outperforms conventional approaches, especially in isolating individual and overlapping nuclei.

\end{abstract}


\IEEEpeerreviewmaketitle

\section{Introduction} 


\IEEEPARstart{F}{rom} its origin in gross anatomy prior to cell theory, to modern computer-aided digital pathology, the field of histopathology has played a critical role in the understanding, diagnosis, and treatment of nearly every disease discovered \cite{Shostak2013,Katz2008,Wright2013}. Millions of tissue biopsies are performed annually, and in nearly every case the study of nuclear morphology and distribution provides critical clues to healthcare providers \cite{Dey2010}. This is not surprising considering the paramount importance of the nucleus, which containing vast amounts of genetic and epigenetic code that govern and regulate cellular type, morphology, and function. Decades of modern cytological and histopathologic study have led to the development of cellular stains such as hematoxylin that stains nuclei \cite{Titford2005,Kapuscinski1979}. For more than a century, the interpretation and determination of aberrant phenotypes in these stained tissue specimens has been accomplished by subjective human interpretation \cite{Titford2012,Alturkistani2016,bandi2018detection}. For the field of computational pathology to mature and impact clinical histology, there is a critical need for accurate, precise and computationally efficient nuclei segmentation methods. \par


The shape and distribution of cell nuclei in pathology images is used to determine cell, tissue, and cancer types, and is critical in cancer identification, grading, and prognosis \cite{Gurcan2009}. For example, in blood smears, multi-lobed nuclei typically indicate neutrophils, oval and kidney-shaped nuclei mark monocytes, round nuclei with a high nuclear-to-cytoplasmic ratio mark lymphocytes, while fully developed red blood cells lack nuclei all together \cite{Lynch1990,Campbell2015}. “Cigar-shaped” nuclei help identify smooth muscle, and, when found out of place, leiomyosarcoma \cite{Katz2008}. Nuclear distribution, morphology, presence, and absence helps identify the layers of epithelium on the exterior and interior of the body, while the identification of cytological penetration of characteristic epithelial layers by cancer such as melanoma can drastically alter treatment course and prognosis \cite{Gershenwald2017}. In breast cancer, which represents one-third of new cancer diagnoses in women, the identification of nuclear and chromosomal mitotic figures provides an accurate prognostic indicator (Elston grading) \cite{Volpi2004,Varma2002}. To reduce prostate cancer mortality, over one million prostate biopsies are performed in the United States annually, and for each, the histopathologist looks for signs of adenocarcinoma by screening for nuclear enlargement and prominent nucleoli \cite{stewart2013clinical}. The importance of nuclear identification and subsequent detailed analysis can hardly be overstated.\par


The field of digital pathology is poised to revolutionize modern healthcare. The advent of whole-slide imagers coupled with telemedicine and cloud storage has facilitated efficient storage of vast amounts of tissue data and promises to facilitate diagnosis, prognosis and treatment planning \cite{Williams2010}. Combining tissue data with electronic medical records, genetic and epigenetic sequencing, big data science, and epidemiologic studies, can enable personalized healthcare and reveal interesting scientific discoveries \cite{Sucaet2014,Louis2016}. The application of precise computer vision techniques to these data is catalyzing the field’s development, increasing the efficiency of providers through rapid screening, aiding in education, and standardizing analysis to reduce observer variability \cite{Gurcan2009,Pantanowitz2012,Stoler2015,Llewellyn2000}. \par

Given the importance of nuclear distribution and morphology the task of precise nuclear segmentation via computer algorithms provides a logical starting point for the rest of computer-aided tissue image analysis. Accurate segmentation of nuclei is a pivotal starting point for further feature extraction and classification within the nucleus itself, but also serves as a relatively simple basis to model cellular distribution, which can be used to classify tissue subtypes and identify abnormalities. \par

While the identification of most nuclei in a conventional H\&E stain is routine for trained clinicians and researchers, chromatic stain variability, nuclear overlap and occlusion, variability in optical image quality, and differences in nuclear and cytoplasmic morphology and stain density provide challenges for computer-based segmentation algorithms \cite{Irshad2014,Xing2016,Xing2016a}. Many techniques have been applied to this task, but have achieved limited success in challenging of cases.  For example, intensity thresholding methods generally fail with noisy images and clumped nuclei, marker-based watershed segmentation requires precise parameter selection, and active contour and deformable models are computationally expensive \cite{Xing2016,Xing2016a,Yang2006,Xue2011,Veta2013,Vahadane2013, Zhang2013}. Machine learning-based nuclear segmentation approaches are generally more robust to these challenges, as they can learn to identify variations in nuclear morphology and staining patterns. More specifically, convolutional neural networks (CNNs) have recently demonstrated state-of-the-art performance in nuclei segmentation \cite{Sirinukunwattana2016,Xing2016a,Kumar2017}. In this approach, images are passed through a trained classifier to label pixels as nuclear or non-nuclear, and additional post-processing techniques to delineate clustered nuclei are subsequently applied \cite{Xing2016a}. A third class of inter-nuclear pixels can be added to eliminate the bulk of the post-processing work, which demonstrates significant improvement in separating crowded nuclei \cite{Kumar2017}. Most current deep learning-based multi-organ nuclei segmentation methods are limited by the amount of data available. For a single network to perfectly segment nuclei from different organs large, diverse, quality-annotated training data is required. Moreover, existing methods often fail to segment overlapping nuclei without complex post processing steps. Lastly, with all of these approaches, the performance is ultimately limited by the accuracy of the training data, which is conventionally bound by human limitations.\par

\textit{\textbf{Contributions:}} In this work, we propose a method to overcome the diversity required in training data using synthetically generated data and then use a contex-aware adversarial network for nuclei segmentation. Our main contributions are highlighted below:
\begin{itemize}
\item \textbf{Synthetic Pathology Images with Ground Truth:} Because of the limited availability of labeled nuclei segmentation data, we generate a large dataset of perfectly-annotated synthetic histopathology images by generating random polygon masks and adding realism using unpaired cycle-consistent adversarial training.
\item  \textbf{Adversarial Nuclei Segmentation:} We propose to train a conditional GAN (cGAN) network with spectral normalization and gradient penalty for multi-organ nuclei segmentation. Instead of using post-processing steps which reinforce spatial contiguity in the nuclei segmentation mask we use adversarial term imposes higher-order spatial consistency during the training process. Moreover, we pose the problem as a regression rather than a classification problem where the loss function is also learned during the training process.
\item \textbf{Quantitative Study:} We validate the proposed nuclei segmentation paradigm on publicly available and newly created datasets. A quantitative study demonstrates the cross site, patient and organ adaptability of our proposed method.
\end{itemize}

\section{Related Work}

\subsection{Deep Learning-based Nuclei Segmentation}

Nuclei segmentation in histopathology images has been extensively studied using a variety of deep learning methods. However, there are several challenges associated with effectively using deep learning for this task. Most work has focused on developing nuclei segmentation methods for single organs and specific application without addressing issues such as domain adaptation \cite{naik2008automated}. Histopathology images are diverse due to variations such as organ type, tissue site, and staining protocol. Clumped and chromatin-sparse nuclei are especially difficult to isolate or detect. The close proximity of epithelial cell and the random occurrence of mitotic figures make accurate boundary detection difficult. Most deep learning techniques estimate a probability map based on a two-class problem followed by post-processing. \textit{Neeraj et al.} completed seminal work in multi-organ nuclei segmentation by posing the challenge as a three class problem (CNN-3C)\cite{Kumar2017}. \textit{Cui et al.} developed a fully convolutional network-based nuclei segmentation approach \cite{cui2018deep} and \textit{Naylor et al.} posed the problem as a regression task of estimating the nuclei distance map. The performance of all of these approaches is limited by the size and quality of labeled datasets and the diversity required in the images to model the distribution of relevant tissue features. For a single method to be adaptable, large amounts of diverse data are required. Moreover, simple CNN-models are generally not capable of handling the issue of overlapping nuclei and post-processing or alternate parallel processing is required to determine such overlaps. This is because CNNs minimize a per-pixel loss between the input image and the segmentation mask. Most previous work has posed the nuclei segmentation problem as a classification or as a combined classification and regression problem. We propose to pose this as an adversarial regression-based image-to-image translation problem. Such an approach is more context aware and globally consistent, \textit{i.e.} the loss function is learned taking the entire image into consideration rather than just pixel-wise loss.

  \vspace{-5mm}
\subsection{GANs for Medical Imaging Applications}
 GANs were introduced by \textit{Goodfellow et al.} in \cite{goodfellow2014generative} and have since been used for a variety of medical imaging applications including segmentation \cite{dou2018unsupervised,li2017brain,rezaei2018whole}, detection \cite{alex2017generative,schlegl2017unsupervised}, reconstruction \cite{quan2017compressed,wang20183d,mardani2017deep}, domain adaptation \cite{zhang2018task,mahmood2018unsupervised,madani2018semi,chen2018semantic}. The GAN framework can be seen as a two player \textit{min-max} game where the first player (the generator), is tasked with transforming a random input to a specific distribution, such that the second player (the discriminator) cannot distinguish between the true and synthesized distributions \cite{goodfellow2014generative,kurach2018gan}. GANs have a generative and artistic ability to map random noise to realistic distributions \cite{wang2018high,isola2017image,zhu2017unpaired}. However, for medical imaging tasks it is critical to constrain this artistic ability and thus conditional GANs (cGANs) are more applicable. cGANs have the ability to conditionally control the output of GAN training based on a class or an image. Image-to-image translation tasks \textit{i.e.} situations where the GAN is conditioned by an input image, are more useful for medical imaging than class conditioning. Due to the limited availability of labeled medical imaging data generative adversarial networks (GANs) have been recently been used for synthetic data generation. For example, \textit{Dou et al.} proposed to generate synthetic retinopathy data from retinal masks \cite{dou2018unsupervised}. \textit{Hou et al.} \cite{schlegl2017unsupervised} proposed generating pathology data using nuclei masks where the background and foreground are learned separately. Despite the wide spread use of GANs they are notoriously difficult to train. We use recently proposed GAN stability methods such as spectral normalization \cite{miyato2018spectral} and gradient penalty for improving the stability of cGAN-based image-to-image translation. 
  \vspace{-4mm}
\section{Methods}
  \vspace{-2mm}
\subsection{Datasets}

There are very few nuclei segmentation datasets with pixel-level annotations. Careful annotation of nuclei boundaries is time-consuming, error-prone and may also suffer from subjective interpretation errors. \textit{Irshad et al.} \cite{irshad2014crowdsourcing} showed that there was larger interoberver disparity among pathologists identifying nuclei in H\&E data. Several datasets have been released for nuclei segmentation but often lack annotations of all nuclei in large portions of the data \cite{gelasca2009biosegmentation,janowczyk2016deep}. This makes it difficult to identify false positives and to focus on clumped and overlapping nuclei. The only publicly available dataset with complete nuclei annotation and multi-organ pathology images was presented by \textit{Kumar et al.} in \cite{Kumar2017}. This dataset contains annotations of $30$ $1000\times1000$ pathology images from seven different organs (bladder, colon, stomach, breast, kidney, liver, and prostate). The raw data for this data was sourced from the NIH Cancer Genome Atlas (TCGA). Although, this is a large database it is still not enough to cater for the diversity required to train context aware methods that can determine overlapping nuclei with a high accuracy on object and pixel-level statistics. To increase the number of organs and test data available we complemented this existing dataset with four additional $1000\times1000$ pathology images (breast, prostate, ovary, and esophagus) from the TCGA database labeled by a pathologist at the Johns Hopkins Hospital. Since, this additional data is labeled by a different pathologist it contributes to the overall diversity of the data. Data from five different organs from the first dataset \cite{Kumar2017} along with synthetically generated data were used for training. Synthetically generated data to models a large diversity of nuclei sizes, overlapping morphology and characteristics that can enhance overall network performance. The trained models were evaluated on a combination of images from the first and second datasets and includes evaluation on five organs that were not used for training.

 \vspace{-5mm}
\subsection{Stain Normalization}
 \vspace{-1mm}

Tissue images stained with H\&E contain significant color variation due to differences in manufacturing of stains, staining protocols used and response functions of digital scanners. Differences in stains can be a major issue in cross-site domain adaptability of CNN-based computational pathology methods. This is because CNNs mostly rely on the color and texture of H\&E images for learning cues. Normalizing the images can significantly improve the performance of these methods \cite{khan2014nonlinear}. However, normalization methods developed for conventional vision applications \cite{reinhard2001color} provide limited benefit in computational pathology applications because the two stains can be normalized against each other. Several methods have been presented for normalizing pathology images or adapting networks trained on data from one site to data from other locations \cite{khan2014nonlinear,bejnordi2016stain,janowczyk2017stain,macenko2009method}. \textit{Abhishek et al.} \cite{vahadane2016structure} proposed decomposing the pathology image into stain density maps in an unsupervised manner and then combining these sparse stain density maps with the stain colored basis of a target image. This approach preserves the structure in the source image while adapting the color to the target domain. We empirically found that sparse stain normalized H\&E images following \cite{vahadane2016structure} performed better with our nuclei segmentation methods as compared to standard \textit{Reinhard} \cite{reinhard2001color} and \textit{Macenko} \cite{macenko2009method} stain normalization. Details about this method can be found in \cite{vahadane2016structure} and examples of stain normalized images have been shown in Fig. 1a.

\begin{figure*}
\centering
\includegraphics[width=\textwidth]{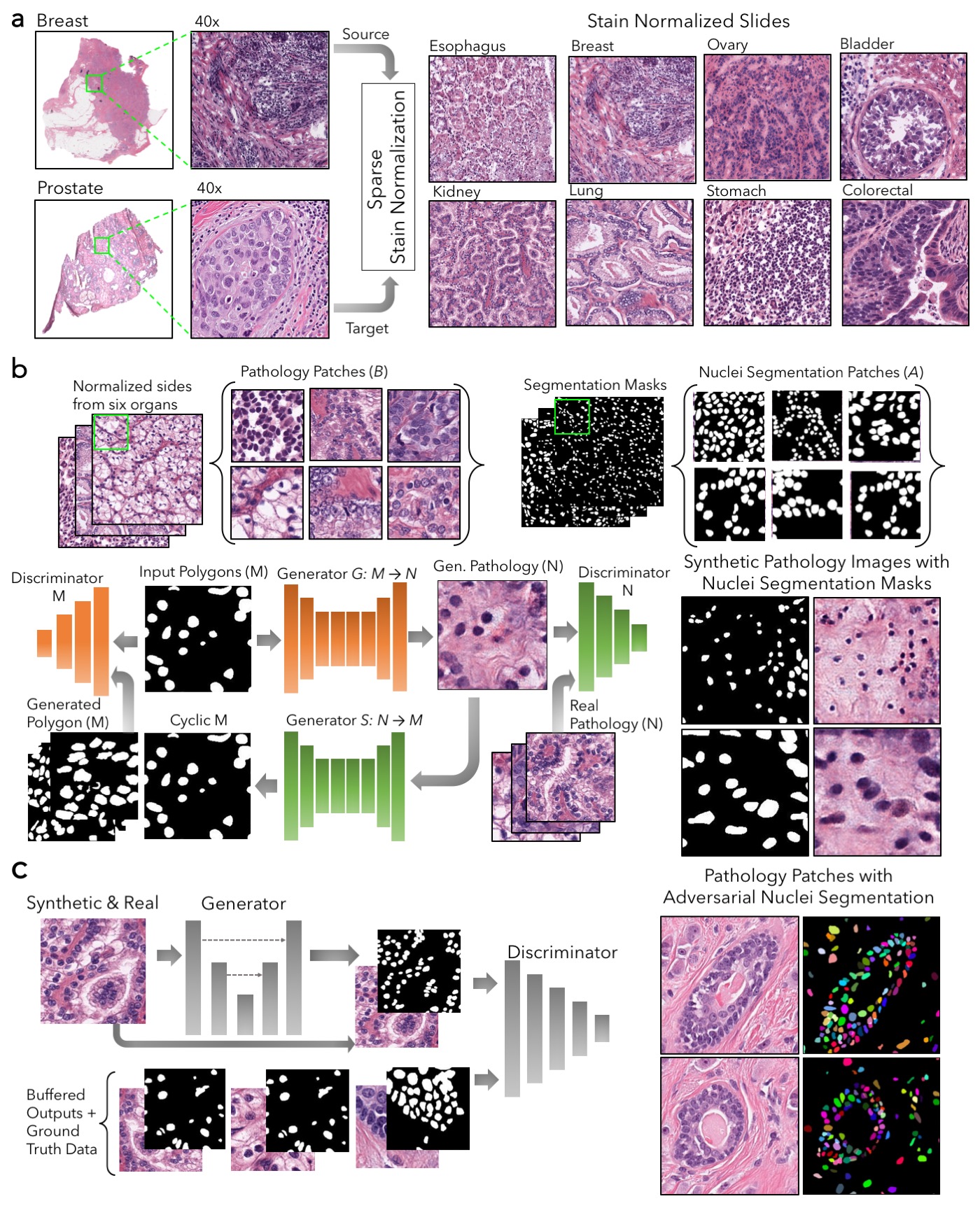}
\caption{(a) Representative full-slide images, $1000\times 1000$ cropped images and sparse stain normalization for nine different organs sources from the NIH TCGA database. (b) Unpaired synthetic data generation using randomly-generated polygon masks mapped to H\&E images. The architecture includes a dual-GAN setup with cycle consistency loss. Two generators learn mappings, $G$, and $S$ between a mask ($M$) and a histology image ($N$) $G: M\rightarrow N$ and $S:N\rightarrow M$ and two discriminators classify the pairs of $M$ and $N$ as real or fake. (c) The conditional GAN setup for segmenting nuclei. The discriminator enhances the receptive field of the generator while learning an efficient loss function for the task.}
\end{figure*}
 \vspace{-2mm}
\subsection{Learning Preliminaries}
 \vspace{-1mm}
\noindent The primary objective of this work is twofold: 

\begin{enumerate}

\item Generate synthetic H\&E images that model the distribution of cellular and extracellular spatial features represented in multiple organs. 
\item Use both the synthetic and real histopathology data for training a context-aware CNN that can accurately segment nuclei. 

\end{enumerate}


This approach can be generalized as learning mapping functions between two domains: $M$ (nuclei mask) and $N$ (H\&E images). Specifically, $G$ maps from $M \rightarrow N$, and is used for synthetic data generation, while $S$ maps from $N \rightarrow N$, and is used for nuclei segmentation. We denote $m$ and $n$ as training examples where $m \in M$ and $n \in N$.


 \vspace{-2mm}
\subsection{Synthetic Data Generation}
 \vspace{-1mm}

Previous work on histopathology image synthesis has focused on generating the nuclei-free background and foreground separately \cite{hou2017unsupervised}. In contrast we propose a relatively simple unpaired mapping-based approach, where cycle consistency loss is used in a dual-GAN architecture to transfer between polygon masks ($M$) and histopathology images ($N$). The size, location and shape of the nuclei can vary significantly based on patients, clinical condition, organs, cell-cycle phase and aberrant phenotypes. Thus, we generate nuclei masks by building a dictionary of nuclei sizes and shapes from different organs and randomly perturbing size and shape parameters before placing them on a grid in a randomized fashion. We then use a cycleGAN-style \cite{zhu2017unpaired}  architecture to add realism to the polygon mask. These generated polygon masks and corresponding synthesized H\&E images are subsequently used for training.  


The cycleGAN framework learns a mapping between randomly generated polygon masks and unpaired pathology images. Since cycleGAN is based on consistency loss, the setup also learns a reverse mapping from pathology images to corresponding segmentation or polygon masks (Fig. 1b). However, since this is an unpaired mapping, the segmentation network is not efficient enough to be used. Thus, we use the reverse mapping only to train the forward mapping more effectively. Such an arrangement consists of four networks: $G$ (random polygon mask to pathology image generator), $S$ (pathology image to polygon mask generator), $D_N$ (discriminator for $G$), and $D_M$ (discriminator for S). To train this framework for synthetic data generation with unpaired data, the cycleGAN objective consists of an adversarial loss term $\mathcal{L}_{\text{GAN}}$ and a cycle consistency loss term $\mathcal{L}_{\text{cyc}}$. The adversarial loss is used to match the distribution of translated samples to that of the target distribution and can be expressed for both mapping functions. The cycle consistency loss term penalizes deviation from the source image. For the mapping $G: M \rightarrow N$ with discriminator $D_N$, we can express the objective as the binary cross entropy (BCE) loss of $D_N$ in classifying real or fake, in which $D_N$ and $G$ play a \textit{min-max} game to maximize and minimize this loss term respectively. This objective for $G: M \rightarrow N$, can be expressed as,


\begin{equation}
  \begin{aligned}
\mathcal{L}_{\text{GAN}}(G, D_{N}) = \mathbb{E}_{n\sim p_{\text{data}}(n)}[ \log D_{N}(n)]\\+\mathbb{E}_{m\sim p_{\text{data}}(m)}[\log(1-D_{N}(G(m)))],
   \end{aligned}  
 \end{equation}

in which the generator $G$ aims to generate nuclei images from random polygon masks such that they would be indistinguishable from real nuclei images, \textit{i.e.}, $G(m) \approx n$, while the discriminator $D_N$ aims to distinguish generated \textit{vs.} real nuclei images. A similar objective can be expressed for $S: N \rightarrow M$,
\begin{equation}
  \begin{aligned}
\mathcal{L}_{\text{GAN}}(S, D_{M}) = \mathbb{E}_{m\sim p_{\text{data}}(m)}[ \log D_{M}(m)]\\+\mathbb{E}_{n\sim p_{\text{data}}(n)}[\log(1-D_{M}(S(n)))].
   \end{aligned}  
 \end{equation}

The cycle consistency loss is used to incentivize a one-to-one mapping between samples in $M$ and $N$, and facilitate the evolution of $G$ and $S$ to inverse functions of each other. Specifically, the $\mathcal{L}_{\text{cyc}}$ term ensures that the forward and back translations between the random polygon mask and nuclei image are lossless and cycle consistent, \textit{i.e.}, $S(G(m)) \approx m$ (forward cycle consistency) and $G(S(N)) \approx N$ (backwards cycle consistency). The forward cycle loss term helps with refining synthetic nuclei images to be more realistic, as the generated images should not only mimic the H\&E tissue, but also the morphology of the nuclei for segmentation. We can express the objective as,


\begin{equation}
  \begin{aligned}
\mathcal{L}_{cyc}(G,S) = \lambda_n \mathbb{E}_{n\sim p_{\text{data}}(n)}[||G(S(n)) - n||_1] \\+ \lambda_m\mathbb{E}_{m\sim p_{\text{data}}(m)}[||S(G(m)) - m||_1]
   \end{aligned}  
 \end{equation}

where $\lambda$ controls the importance of the forward and backward cycle constraints. For synthetic data generation, we relaxed the $\lambda_m$ term, as a random polygon mask can represent multiple valid nuclei images. This randomization can also be seen as GAN input noise that contributes to generating the diversity in synthetic image.

The full objective for synthetic data generation can thus be written as,

 \vspace{-5mm}
\begin{equation}
  \begin{aligned}
\argminA_{G,S} \argmaxA_{D_N, D_M} \mathcal{L}_{\text{GAN}}(G, D_{N}) + \mathcal{L}_{\text{GAN}}(S, D_{M}) + \mathcal{L}_{cyc}(G,S)
   \end{aligned}  
 \end{equation}
 \vspace{-10mm}
\subsection{Conditional GANs for Segmentation}
 \vspace{-1mm}
One of the major challenges in nuclei segmentation is independent boundary detection and isolation of overlapping nuclei. Standard CNNs segment these nuclei as one object because these approaches typically rely on minimization of some pixel-wise loss. The contribution of a single misclassified pixel is insignificant to the overall loss but can subsequently lead to multiple nuclei segmented as one. This problem has previously been mitigated by contour prediction \cite{Kumar2017,van2016deep,tareef2018multi}, concave point detection \cite{zhang2017automated} and distance map regression \cite{naylor2018segmentation}.

For conventional vision, conditional random fields (CRFs) have been extensively used to enforce spatial contiguity as a post-processing step in segmentation problems \cite{chen2018deeplab}. Joint CNN-CRF models have also been explored for more global and context aware CNN training \cite{mahmood2018unsupervised,mahmood2018deep,mahmood2018deepcine}. Despite the advances in CNN-CRF models, this approach is often limited to the use of pairwise CRFs, which only incorporate second order statistics. According to \cite{luc2016semantic,chen2018rethinking} higher order potentials have also been useful for image segmentation. Using CRFs to incorporate higher order statistics renders a complex energy function. Adversarial training allows higher order consistency without being limited to a specific type of higher order potential (\textit{e.g.} unary and pairwise in the case of CRFs). Since the adversarial model has a field-of-view that is a large portion of the image rather than just neighboring pixels or super-pixels, it can enforce a higher-order statistical consistency that can neither be enforced using pair-wise terms in a CRF nor measured by pixel-wise loss.

The adversarial segmentation model also learns an appropriate loss function which circumvents manually engineered loss functions. This has been explored in detail in image-to-image translation methods such as \cite{isola2017image}. Such a model is flexible enough to detect subtle differences in a range of higher order statistics between the predicted and ground truth nuclei segmentation masks.  The adversarial setup can learn a loss, based on classifying the output image as real or fake, while iteratively training a segmentation model to minimize this learned loss. Each output pixel is usually considered conditionally independent from all other pixels whereas conditional GANs can learn a structured context-aware loss considering a larger receptive field.

The cGAN framework learns a mapping $S$ for nuclei segmentation, in which $S$ can adapt H\&E nuclei images to their segmentation masks. To train this framework for semantic segmentation with paired data, the conditional GAN objective consists of an adversarial loss term $\mathcal{L}_{\text{GAN}}$ and a per-pixel loss term  $\mathcal{L}_{1}$ to penalize both the joint configuration of pixels and segmentation errors.

The adversarial loss in conditional GANs is similar to that of cycleGAN, in which the segmentation network $S$ and discriminator $D_{M}$ play a \textit{min-max} game in respectively minimizing and maximizing the objective, $\text{min}_S \text{max}_{D_M} \mathcal{L}_{\text{GAN}}(S, D_{M})$. Specifically, $S$ translates nuclei images to realistic segmentation masks to minimize the cross-entropy loss of $D_M$. The adversarial loss can additionally be interpreted as a structured loss, in which $S$ is penalized if the group configuration of the pixels in the predicted mask is unrealistic, \textit{i.e.} masks that look like salt-and-pepper noise. Because the data is paired, $D_M$ sees both the nuclei image and the predicted mask. We can express the GAN objective as,

\begin{equation}
  \begin{aligned}
\mathcal{L}_{\text{GAN}}(S, D_{M}) = \mathbb{E}_{m,n\sim p_{\text{data}}(m,n)}[ \text{log}D_{M}(m,n)]\\+\mathbb{E}_{n\sim p_{\text{data}}(n)}[\text{log}(1-D_{M}(m,S(n)))]
   \end{aligned}  
 \end{equation}

\noindent An additional  $\mathcal{L}_1$ loss term is used to bring the output closer to the ground truth and stabilize GAN training,

$$ \mathcal{L}_{1}(S) = \mathbb{E}_{m,n\sim p_{\text{data}}}(m,n)[||m-S(n)||_1]. $$

\noindent The full objective for conditional GAN-based segmentation can be expressed as, 
\begin{equation}
  \vspace{-2mm}
  \begin{aligned}
  \vspace{-2mm}
 \text{arg min}_S \text{max}_{D_M} \mathcal{L}_{\text{GAN}}(S, D_{M}) +  \mathcal{L}_{1}(S). 
   \end{aligned}  
 \end{equation}
Because the discriminator works on overlapping patches (\textit{i.e.,} PatchGAN Markovian classifier), it penalizes structure at a patch level rather than over the entire image. This approach draws focus of the network on portions of the image in which the nuclei boundaries are likely to be missed. The overlapping patches mean that the same nuclei of the image contribute to the learned loss multiple times in a different context and varying neighboring environments. 

\subsection{Spectral Normalization for GAN Stability}

Discriminator normalization can improve stability. From an optimization point-of-view, such normalization leads to more efficient gradient flow. Various forms of normalization have been proposed, for example batch normalization (BN) was proposed for GAN frameworks in \cite{denton2015deep}. BN is done on a batch level, and normalizes pre-activations of nodes in a layer to the mean and standard deviation of the parameters learned for each node in the layer. Since a neural network can be seen as a composition of non-linear mappings with spectral properties. Spectral normalization was first suggested in \cite{miyato2018spectral} for improving GAN stability and entails dividing each weight matrix, including the matrix representing convolutional kernels, by their spectral norm. We use spectral normalization for stabilizing GAN training for both synthetic data generation and segmentation.

\section{Experiments and Results}

\subsection{Implementation Details}

\noindent \textbf{Dataset Preprocessing}

\noindent The data obtained from \cite{Kumar2017} and our manually labeled data were normalized using sparse stain normalization to match a standard slide from the breast. Collectively, these data were sourced from nine different organs, 34 different patients and collected at different hospitals. Four slides from breast, liver, kidney and prostate were used for training. For testing, two slides from breast, liver, kidney, prostate, bladder, colon and stomach and one slide from esophagus and ovary were used. Thus, our test data includes images from five organs that the network was not trained on. All data were decomposed into large patches of size $256\times 256$ for training the network efficiently. The training data was then supplemented with synthetically generated data containing $4,650$ patches. The overall training data had $4906$ $256\times 256$ patches. 

\noindent \textbf{Synthetic Data Generation}

\noindent \textbf {\textit{Network Architectures:}} The generator architectures contain two stride-2 convolutions, nine residual blocks and two functionally constrained convolutions with a stride of  $\frac{1}{2}$. Reflection padding was used to minimize artifacts. The discriminator architecture was a simple classifier with three layers and the output was $70 \times 70$ with the aim to classify weather these overlapping patches were real or fake. As suggested in \cite{zhu2017unpaired} a patch level discriminator has fewer parameters and is more easily applicable to various image sizes.  We observed that larger size images needed more residual blocks for efficient convergence. The GAN training was stabilized to prevent mode collapse by using spectral normalization \cite{miyato2018spectral}. 

\noindent \textbf {\textit{Training Details:}} The training code was implemented using Pytorch 0.4. For all experiments $\lambda_n = 70$ and $\lambda_m = 10$. Adam solver \cite{kingma2014adam} was used to solve the optimization problem with a batch size of 1, which was experimentally determined. A total of $300$ epochs was used. The learning rate was set to $0.0002$ for the first $150$ epochs and linearly decayed to zero for the remaining $150$ epochs. Since the purpose of this architecture was to construct an accurate generator, we divided the objective function by two when optimizing the discriminator \textit{i.e.} to give it a lower learning rate. All networks were trained from scratch with no prior knowledge and weights were initialized from a Gaussian distribution with a mean and standard deviation of $0$ and $0.02$ respectively.  

\noindent \textbf{Nuclei Segmentation}

\noindent \textbf {\textit{Network Architectures:}} We use an encoder-decoder architecture with skip connections (U-Net \cite{ronneberger2015u}) for the generator. Skip connections are added between the $i^{th}$ layer and the $(n-i)^{th}$ layer, where each skip concatenates channels at the $i^{th}$ and the $(n-i)^{th}$ layers, where $n$ is the total number of layers. After the last layer in the decoder, a convolution is applied to map the output segmentation mask followed by a \textit{Tanh} function. Leaky ReLUs \cite{xu2015empirical} were employed for the encoder with a slope of 0.2 and regular ReLUs were used for the decoder. A $70 \times 70$ patch Markovian discriminator, similar to the one described previously \cite{isola2017image}, was employed but with leaky ReLUs with a slope of 0.2. 

\noindent \textbf {\textit{Training Details:}} During training, random jitter was applied to the data by resizing the $256 \time 256$ patches to $286\times 286$ and then cropping them back to $256\times 256$. The Adam optimizer was used to solve the objective function and all networks were trained from scratch where the weights were initialized from a Gaussian distribution of mean 0 and standard deviation 0.02. The training was run for $400$ epochs and the learning rate was set to $0.0002$ for the first $200$ epochs and linearly decayed to zero for the remaining $200$ epochs. Spectral normalization was used for increasing the stability of the adversarial training. A pooling history of randomly selected $64$ patch pairs from the segmented output and ground truth was used in the discriminator.  

\begin{table*}
 \caption{Comparative Analysis of the proposed Nuclei Segmentation Framework Against Standard Architectures from the test dataset}
 \vspace{-5mm}
\begin{center}
\begin{tabular}{|c|c|c|c|c|c|c|c|c|c|c|c|c|c|}
\hline
\multirow{2}{*}{Organ} & \multicolumn{4}{c|}{\textbf{Aggregated Jaccard Index (AJI) $\uparrow$  }} & %
    \multicolumn{4}{c|}{\textbf{Average Hausdorff distance $\downarrow$}} & \multicolumn{4}{c|}{\textbf{$\textbf{F1}$-Score $\uparrow$}} \\
\cline{2-13}
 & FCN & U-Net & Mask R-CNN & \textbf{Proposed} & FCN & U-Net & Mask R-CNN & \textbf{Proposed} & FCN & U-Net & Mask R-CNN & \textbf{Proposed}\\
\hline
 \textbf{Breast} & 0.339& 0.485& 0.483& 0.686 & 8.17& 7.927& 6.286&  4.761&0.794&  0.774& 0.749&0.881\\
\hline
 \textbf{Ovary} & 0.322&  0.471& 0.513& 0.728 & 8.48& 8.142& 7.236&4.221&0.675&  0.702& 0.725&0.876\\
\hline
  \textbf{Esophagus} & 0.293& 0.392& 0.433& 0.759& 8.71&  8.349&  7.394&3.963&0.667& 0.698&0.733&0.891\\
\hline
  \textbf{Liver} & 0.364 & 0.422& 0.469 & 0.692& 7.93& 7.463& 6.927&  4.173&0.683&  0.697&0.741&0.794\\
\hline
  \textbf{Kidney} & 0.472 & 0.532& 0.542& 0.724& 7.34&  6.858&  6.538&4.492&0.775&  0.783&0.779&0.829\\
\hline
  \textbf{Prostate} & 0.281& 0.276& 0.536& 0.731& 8.94& 8.347& 7.912&  4.013&0.794& 0.802&0.784& 0.857\\
\hline
  \textbf{Bladder} & 0.461& 0.491& 0.477& 0.768& 7.11& 7.423& 7.612& 4.273 &0.746& 0.791&0.769& 0.904\\
\hline
  \textbf{Colorectal} & 0.247& 0.218& 0.391& 0.686& 9.27&  8.729& 7.743& 4.712&0.719& 0.733&0.691& 0.836\\
\hline
  \textbf{Stomach} & 0.383& 0.437& 0.629& 0.721& 7.52& 7.314& 7.518&  4.017&0.866& 0.874&0.843& 0.922\\
\hline
\hline
  \textbf{Overall} & 0.351& 0.414& 0.497& \textbf{0.721}& 8.163& 7.839 & 7.136& \textbf{4.291}& 0.746& 0.761& 0.757&\textbf{0.866}\\
\hline
\end{tabular}
\end{center}
\vspace{-4mm}

\end{table*}

\vspace{-2mm}
\subsection{Evaluation Criteria}

For efficient evaluation of nuclei segmentation, both object-level and pixel-level errors must be penalized. Object-level penalty is associated with nuclei detection and pixel-level penalty is important for preserving fine boundaries, which influence the shape and size of the nuclei mask. As described in \cite{Kumar2017} any quantitative nuclei segmentation metric should penalize four different possible erroneous outcomes: a) Missed nuclei b) Detection of ghost objects as nuclei c) Under-segmentation (\textit{i.e.,} segmentation of overlapping nuclei as one) d) Over-segmentation (\textit{i.e.,} segmentation of a single nucleus as many). For our quantitative study we use three different evaluation metrics that assess these outcomes:  

\begin{itemize}

\item \textbf{Average Pompeiu\textendash Hausdorff distance (aHD):} The aHD is a pixel-level metric that calculates the greatest of all the distances from a point in the ground truth segmentation mask to the closest point in the predicted mask \cite{huttenlocher1993comparing}. It can be calculated as $P(x,y)=\max(\vec{p}(h,y),\vec{p}(y,h))$ where $\vec{p}=\max_{a} \min_{b}\norm{a-b}$ and $a\in h$ and $b\in y$. A lower HD indicates the two segmentation masks being compared are closer to each other.
\item \textbf{$F_1$ Score:} The $F_1$ Score is an object-level metric defined by the harmonic mean between precision and recall, $F_1=\frac{Recall\times Precision}{Recall+Precision}$. A higher $F_1$ score indicates a better intersection between the ground truth and the predicted segmentation masks.
\item \textbf{Aggregated Jaccard Index (AJI):} AJI was proposed in \cite{Kumar2017} specifically for quantitative evaluation of nuclei segmentation. AJI is an extension of the global Jaccard index which computes the ratio of aggregated intersection cardinality and aggregated union cardinality in the ROI. Assuming, $G=\bigcup_{i=1,2..K}G_i$ is the ground truth of the nuclei pixels and $P=\bigcup_{j=1,2..L}P_j$ are the prediction results, the AJI can then be defined as, $AJI=\frac{\sum_{i=1}^L{|G_i\cap P_j^*(i)|}}
{\sum_{i=1}^K {|G_i\cup P_j^*(i)|+\sum_{k\in U}|P_k|}}$. Where, $P_j^*(i)$ is the connected component from the prediction result for which the Jaccard index is maximized, and U is the set of indices of detected ghosts that are not assigned to any component in the ground truth. A higher AJI indicates better results while penalizing all four possible errors mentioned above.

\end{itemize}

\begin{figure*}
\centering
\includegraphics[width=\textwidth]{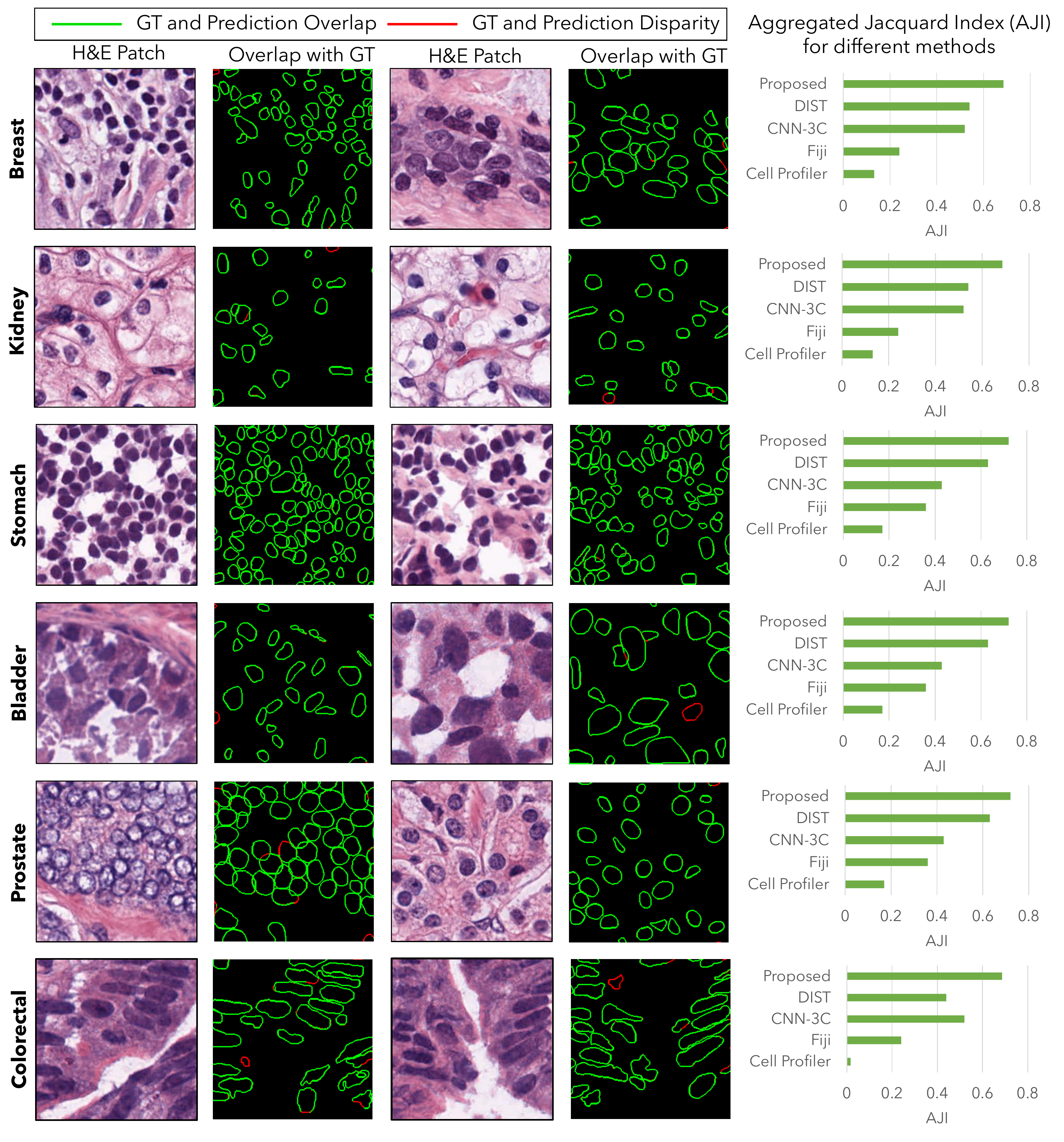}
\caption{Representative patches from six different organs and corresponding nuclei segmentation masks predicted by our proposed method, overlaid on ground truth segmentation masks. The green region represents an overlap between the prediction and manually labeled ground truth whereas the red region represents a disparity between the two. The predominance of the green region demonstrates accurate labeling. The bar charts compare the AJI for all test patches of corresponding organs with state-of-the-art methods (DIST and CNN-3C) as well as commonly-used segmentation tools in Fiji and Cell Profiler.}
\end{figure*}

\vspace{-5mm}
\subsection{Quantitative Study and Results}

We used data from nine different organs for evaluating the trained nuclei segmentation network. No data used for testing was used for training. To assess generalizability of the developed algorithms, we tested on five organs that were not represented in the real or synthetic training data. Table I summarizes the results of the testing on three different metrics mentioned above. A comparative analysis with standard segmentation architectures \cite{garcia2017review,litjens2017survey} such as U-Net \cite{ronneberger2015u}, Fully Convolutional Network (FCN) \cite{chen2018deeplab} and Mask R-CNN \cite{he2017mask} was performed using an identical test and train split. These architectures were used for comparison because of their wide spread usage for a variety of different segmentation tasks including nuclei segmentation \cite{ho2017nuclei,wollmann2018multi}. Besides these standard architectures, we also make explicit comparisons with DIST \cite{naylor2018segmentation} and CNN-3C \cite{Kumar2017} two state-of-the-art nuclei segmentation methods. We also compare with nuclei segmentation toolboxes available in Cell Profiler \cite{carpenter2006cellprofiler} and ImageJ-Fiji \cite{schindelin2012fiji}. These comparisons and representative patches from different organs are shown in Fig. 2. The overlap between ground truth annotations from a pathologist and those detected by our method are highlighted in green and the difference between the ground truth and the prediction are shown in red. Our segmentation networks demonstrates a \textbf{$\textbf{29.19\%}$} improvement in AJI as compared to DIST and \textbf{$\textbf{42.98}\%$} as compared to CNN-3C. In terms of standard architectures there is a \textbf{$\textbf{44.27}\%$} improvement over standard Mask R-CNN and \textbf{$\textbf{73.19}\%$} over a U-Net. 

\vspace{-4mm}
\section{Discussion and Conclusion}

Objective, accurate, and automated analysis of H\&E slides has the potential to revolutionize clinical practice in pathology, by improving the detection, grading, classification, and quantitative analysis of aberrant phenotypes. Towards realizing this potential, this work addresses one of the most fundamental tasks in computational pathology—nuclear segmentation. Nuclear morphology and distribution are paramount for the analysis of histopathology slides by computational pathology. In addition to providing foundational features, nuclei shape and position can be used to enhance network attention. 

We propose a single network that is trained with four organs and synthetically generated pathology data. Our network is trained using an adversarial pipeline which has a larger receptive field as compared to standard CNNs and captures more global information. This approach captures higher-order statistics from the image and the resulting networks are more context-aware. We pose the segmentation problem as an image-to-image translation task rather than a classification task. Doing so allows us to learn a complex loss function between the output and the ground truth rather than having to use a manually engineered one. We demonstrate that this approach performs better than standard architectures, state-of-the-art methods and general purpose tools such as Fiji and Cell Profiler for nuclei segmentation. 

Future work will involve adapting this approach to other medical imaging modalities, as well as fusing nuclear morphology information with other features for networks that can improve detection, classification, grading and prognosis from histopathology images. Future work will also explore the adversarial training paradigm for structured prediction and generalizing the concept for other applications. The nuclei segmentation training code is publicly available at: \textit{github.com/faisalml/NucleiSegmentation}
\vspace{-3mm}
\section*{Acknowledgment}

The authors would like to thank subsidized computing resources from Google Cloud.

\ifCLASSOPTIONcaptionsoff
\fi



\end{document}